\documentclass[conference]{IEEEtran}
\IEEEoverridecommandlockouts
\usepackage{cite}
\usepackage{amsmath,amssymb,amsfonts}
\usepackage{algorithmic}
\usepackage{graphicx}
\usepackage{textcomp}
\usepackage{xcolor}
\usepackage{multirow}
\usepackage{booktabs} 
\usepackage{makecell}
\usepackage{placeins}
\usepackage{float}
\usepackage{bm}
\def\BibTeX{{\rm B\kern-.05em{\sc i\kern-.025em b}\kern-.08em
    T\kern-.1667em\lower.7ex\hbox{E}\kern-.125emX}}
\usepackage{overpic}
\usepackage{adjustbox}
\usepackage{subfig}
\makeatletter
\def\ps@IEEEtitlepagestyle{%
  \def\@oddfoot{\mycopyrightnotice}%
  \def\@evenfoot{}%
}
\def\mycopyrightnotice{%
  {\footnotesize 979-8-3503-9114-5/25/\$31.00 © 2025 IEEE\hfill}%
  \gdef\mycopyrightnotice{}
}
\makeatother

\begin{document}

\title{D-PerceptCT: Deep Perceptual Enhancement for Low-Dose CT Images
}
\author{
\IEEEauthorblockN{Taifour Yousra Nabila}
\IEEEauthorblockA{\textit{Sorbonne Paris Nord University} \\
Villetaneuse, France}
\and
\IEEEauthorblockN{Azeddine Beghdadi}
\IEEEauthorblockA{\textit{Sorbonne Paris Nord University} \\
Villetaneuse, France}
\and
\IEEEauthorblockN{Marie Luong}
\IEEEauthorblockA{\textit{Sorbonne Paris Nord University} \\
Villetaneuse, France}
\and
\IEEEauthorblockN{Zuheng Ming}
\IEEEauthorblockA{\textit{Sorbonne Paris Nord University} \\
Villetaneuse, France}
\and
\IEEEauthorblockN{Habib Zaidi}
\IEEEauthorblockA{\textit{Geneva University Hospital} \\
Geneva, Switzerland}
\and
\IEEEauthorblockN{Faouzi Alaya Cheikh}
\IEEEauthorblockA{\textit{NTNU University} \\
Gjovik, Norway}
}

\maketitle
\begin{abstract}
Low Dose Computed Tomography (LDCT) is widely used as an imaging solution to aid diagnosis and other clinical tasks. However, this comes at the price of a deterioration in image quality due to the low dose of radiation used to reduce the risk of secondary cancer development.
While some efficient methods have been proposed to enhance LDCT quality, many overestimate noise and perform excessive smoothing, leading to a loss of critical details. In this paper, we introduce D-PerceptCT, a novel architecture inspired by key principles of the Human Visual System (HVS) to enhance LDCT images. The objective is to guide the model to enhance or preserve perceptually relevant features, thereby providing radiologists with CT images where critical anatomical structures and fine pathological details are perceptually visible. D-PerceptCT  consists of two main blocks: 1) a Visual Dual-path Extractor (ViDex), which integrates semantic priors from a pretrained DINOv2 model with local spatial features, allowing the network to incorporate semantic-awareness during enhancement;
(2) a Global-Local State-Space block that captures long-range information and multiscale features to preserve the important structures and fine details for diagnosis.
In addition, we propose a novel deep perceptual loss, designated as the Deep Perceptual Relevancy Loss Function (DPRLF), which is inspired by human contrast sensitivity, to further emphasize perceptually important features. Extensive experiments on the Mayo2016 dataset demonstrate the effectiveness of  D-PerceptCT method for LDCT enhancement, showing better preservation of structural and textural information within LDCT images compared to SOTA methods.



\end{abstract}

\section{Introduction}

Computed Tomography (CT) imaging provides detailed insights into fine structures, making it indispensable for accurate diagnosis and radiotherapy planning. However, standard-dose CT scanning raises serious health concerns due to the high X-ray exposure and the associated risk of secondary cancers. Low-dose CT (LDCT) scanning is used as an efficient alternative solution to minimize radiation exposure, but at the expense of deteriorated image quality due to noise and artifacts that reduce the visibility of critical details, such as tumors and lesions. These degradations not only affect expert interpretation but also significantly affects the performance of computer-aided diagnosis systems, such as organ segmentation \cite{salimi2025deep}.

A plethora of methods has been proposed in the literature \cite{kulathilake2023review} for LDCT enhancement. Early CNN-based methods \cite{chen2017low} suppressed noise but tend to over-smooth structures, thereby eliminating fine details and textures crucial for diagnosis. Generative adversarial networks (GANs) \cite{liu2023solving,yang2018low} widely explored to enhance LDCT image quality. Notably, Yang et al. \cite{yang2018low} proposed a GAN-based solution incorporating a deep perceptual loss that helps maintain sharpness and prevent over smoothing. More recently, transformer-based models \cite{wang2023ctformer,chen2024lit} have shown promising results. In particular, Chen et al. \cite{chen2024lit} addressed both denoising and deblurring in enhanced LDCT images. Diffusion models \cite{xia2022low,gao2022cocodiff} and UNet-based architectures \cite{zhang2023novel,xiong2024re} have also been extensively explored, proving highly effective in preserving details and texture information.

Recently, state-space models have achieved remarkable results, outperforming the previously discussed methods. Approaches such as CTMamba \cite{li2024ct} and DenoMamba \cite{ozturk2024denomamba} have demonstrated significant improvements in LDCT image enhancement.

Despite significant advancements in LDCT image denoising, existing methods often over smooth small structures, textures, and relevant fine details, thereby compromising the diagnosis accuracy. In this work, we propose a solution based on visual perception approach and therefore integrating the dimension of the human observer. This approach consists in integrating into the proposed architecture both low and high-level aspects of the Human Visual System (HVS), namely: prior semantic understanding; multiscale visual information processing; global-local attention mechanisms; and the HVS’s varying sensitivity to different spatial frequency bands \cite{shapley1985spatial}. Therefore, we propose D-PerceptCT, a perceptual LDCT enhancement framework that integrates these aspects at multiple levels of the architecture to improve the perceptual quality of LDCT images.



In summary, the key contributions of this works are as
follows:
\begin{itemize}
\item We propose D-PerceptCT, a novel LDCT enhancement framework that  integrates human visual perception mechanisms such as multiscale processing, global-local visual information processing, semantic awareness, and visual contrast sensitivity with advanced architectures including DINOv2 and Vision State-Space modeling, to enhance the visibility of anatomical structures and fine details in low-dose CT images.

\item We employ pretrained DINOv2 \cite{oquab2024dinov} features as semantic priors within a dual-branch architecture, enabling the network to incorporate high-level visual context during enhancement.

\item We introduce a novel Deep Perceptual Relevance Loss Function (DPRLF), which incorporates the contrast sensitivity of the HVS and leverages multi-resolution analysis to effectively capture the perceptual relevance of image features.

\item We conducted extensive evaluations using both  conventional image quality metrics (e.g., PSNR, SSIM) and perceptual quality measures (e.g., LPIPS, DISTS, PIQE) that better reflect perceptual similarity. Additionally, we perform pairwise comparisons and compute an accumulated score to establish a comprehensive ranking among competing methods.
\end{itemize}

The rest of this paper is organized as follows: Section II presents the proposed D-PerceptCT framework. Section III describes the dataset, implementation details, and evaluation metrics used in our experiments. Section IV discusses the quantitative and qualitative results, along with comparative and ablation studies. Finally, Section V concludes the paper and outlines directions for future work.
\section{The proposed method}
In the following, we provide an overview of the proposed
D-PerceptCT method for LDCT image quality enhancement while focusing on the most relevant and original elements developed in this study.
\subsection{Visual Dual-Path Extractor Module (ViDex)}

Given a degraded LDCT image $\bm{
I} \in \mathbb{R}^{1 \times H \times W}$, we first extract relevant features with a biologically inspired dual-branch Module. We draw inspiration from previous studies on HVS  revealed
that two visual pathways, namely the dorsal and ventral pathways, process visual information simultaneously in the visual cortex. The ventral stream, often referred to as the “what” pathway, involves semantic information extraction to achieve a high-level  understanding of structures and objects, while the dorsal stream, or “where/how” pathway processes spatial information and captures  fine visual details. Prior works leveraging dual-pathway architectures have demonstrated their effectiveness in tasks, including image quality assessment and edge detection~\cite{chen2023using, chen2022dped}. 

We incorporate these dual streams within our Visual Dual-path Extractor Module (ViDex), which consists of two branches, namely Semantic Feature Extractor Branch (SFEB) and Local detail extractor (LDEB), as illustrated in Figure \ref{fig:DPVE} below. 

\subsubsection{Semantic Feature Extractor Branch (SFEB)}
The Semantic Feature Extractor Branch (SFEB) mimics the role of the ventral stream in the HVS, which is responsible for semantic understanding and high-level perception. In the context of LDCT enhancement, semantic awareness refers to the model's ability to identify and preserve relevant structures during the denoising process. To this end, SFEB extracts semantic priors using a pre-trained DINOv2 vision transformer~\cite{oquab2024dinov}, which has shown remarkable performance in various visual tasks such as classification, segmentation, and retrieval~\cite{huang2024dino,song2024dino,jose2025dinov2}.


\begin{figure}[htbp]
    \centering
    \includegraphics[scale=0.68]{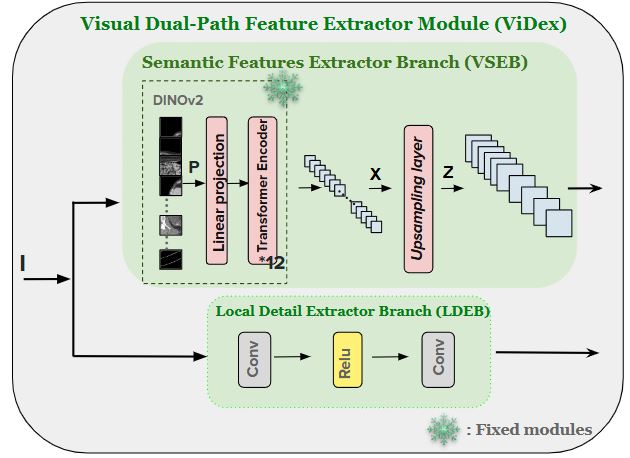} 
    \caption{Architectural Diagram of the proposed  Visual Dual-path Extractor Module (ViDex).}
    \label{fig:DPVE}
\end{figure}
\begin{figure}[htbp]
    \centering
    \includegraphics[scale=0.5]{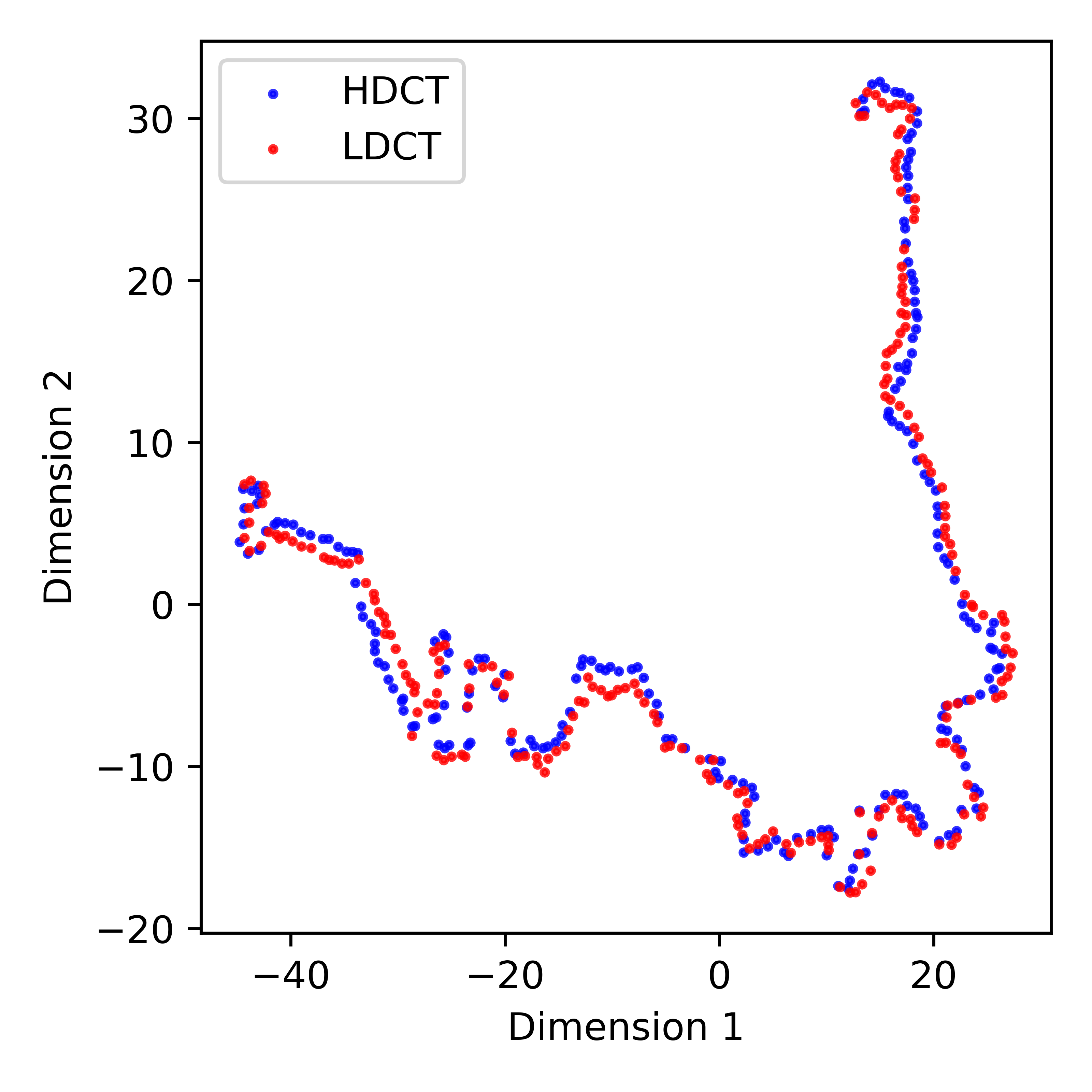} 
    \caption{Visualization of DINOv2 embeddings via t-SNE for paired low-dose (red) and high-dose (blue) CT slices of patient L506 from the Mayo2016 dataset.}
    \label{fig:dino}
\end{figure}

Given an input LDCT image $I$, it is first partitioned into non-overlapping patch tokens $P \in \mathbb{R}^{16 \times 16 \times 3}$. These tokens are then passed through a twelve consecutive transformer encoders to produce high-level semantic embeddings $X \in \mathbb{R}^{C \times H' \times W'}$, where $C = 192$, and $H' = \frac{H}{16}$, $W' = \frac{W}{16}$. These embeddings $X$ are subsequently upsampled to generate a dense semantic feature map $Z \in \mathbb{R}^{C \times H \times W}$, restoring the original spatial resolution and enabling pixel-wise semantic guidance. This enhanced representation guides the model in suppressing noise while preserving important anatomical structures and edges in the LDCT image.
We also conducted a preliminary experiment across all patients in the Mayo2016 dataset \cite{mccollough2016tu} to assess the consistency of DINOv2's semantic representations across low-dose (LDCT) and high-dose CT (HDCT) scans. Figure \ref{fig:dino} shows a representative example from patient L506, t-SNE visualization of the extracted embeddings from each LDCT (red) and HDCT images (blue) projected on two dimensions, demonstrates a strong correlation, indicating that DINOv2 extracts dose-invariant semantic features. The integration of such information promotes structure-preserving enhancement of LDCT images. 

Moreover, the patch-wise processing of DINOv2 mirrors the region-by-region diagnostic reasoning employed by radiologists, thereby aligning our method closely with human clinical workflows.

\subsubsection{Local Detail Extractor Branch (LDEB)}

In parallel to SFEB, the low-dose CT image \( \mathbf{I} \in \mathbb{R}^{H \times W \times C} \) is passed through a lightweight convolutional model consisting of two sequential convolutional layers separated by a ReLU non-linearity:
\[
F_{\mathrm{LDEB}} = W_{2} * \bigl(\mathrm{ReLU}(W_{1} * \mathbf{I})\bigr),
\]
where \( W_1 \) and \( W_2 \) denote convolution kernels with a spatial size of \(3 \times 3\), and \(*\) represents the convolution operation.

This branch mimics the dorsal stream, which is responsible for processing spatial details and local information such as textures, edges, and noise patterns that are critical for LDCT image enhancement. The resulting feature maps are then passed through a lightweight Feature Fusion Module (2FM), which enables joint learning of global semantic features and local details, producing enriched feature maps that are subsequently fed into the Deep Visual State-Space Model (DV2SM) for refined enhancement.


\subsection{The proposed method}
\begin{figure}[htbp]
    \centering
    \includegraphics[width=1\linewidth]{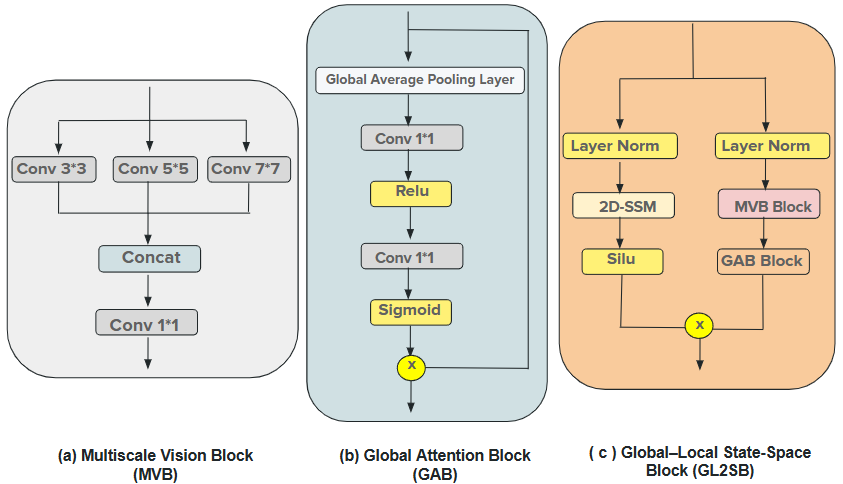}
    \caption{Representative blocks of DV2SM. (a) Multiscale Vision Block, (b) Global Attention Block, 
    (c) Global-Local State Space Block (GL2SB)}
    \label{fig:blocks}
\end{figure}
Our proposed Deep Visual State Space (DV2SM) learning model consists of four Visual State Space Groups (VSSGs), each composed of three novel Global-Local State Space blocks (GL2SB). This architecture builds upon the MambaAIR baseline~\cite{guo2024mambair}, a Mamba-based state-space model developed for natural images reconstruction. 

To effectively enhance the visibility of fine structures often blurred or suppressed in prior works, we propose the Global-Local State Space Block (GL2SB), combining HVS-inspired perception with the effectiveness of Mamba-based architectures. The GL2SB ( Figure \ref{fig:blocks}) is an extension of the previously proposed Visual State Space block introduced in~\cite{guo2024mambair} by incorporating global multiscale visual information processing in the primary visual cortex (V1). 

The architecture of GL2SB consists of two branches joint by a learnable skip scale mechanism:
\begin{itemize}
    \item Spatial self-attention via a 2D state-space block (SS2D) to capture long-range information within the image;
    \item Multiscale Convolutional Attention through a Multiscale Vision Block (MVB), which captures local details and global contextual features across different spatial resolutions. As depicted in Figure~\ref{fig:blocks}, the MVB leverages parallel convolutions with varying kernel sizes to represent anatomical structures of different sizes, reflecting the multiscale spatial sensitivity of V1 neurons in the HVS. Then a channel-wise fusion is performed via concatenation followed by a \(1 \times 1\) projection.
    
\end{itemize}

\begin{figure*}[ht]
    \centering
    \begin{minipage}{\textwidth}
        \subfloat[]{%
            \begin{overpic}[width=0.18\textwidth]{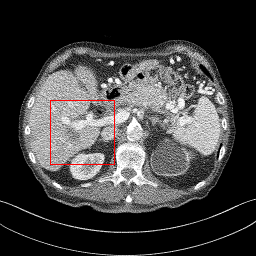}
                \put(55,-6){\adjustbox{frame=1pt,bgcolor=white}{\includegraphics[width=0.12\linewidth]{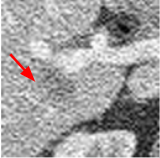}}}
            \end{overpic}
        }\hfill
        \subfloat[]{%
            \begin{overpic}[width=0.18\textwidth]{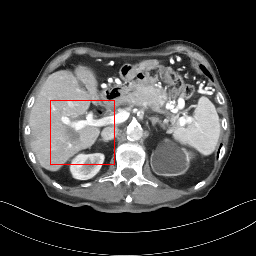}
                \put(55,-6){\adjustbox{frame=1pt,bgcolor=white}{\includegraphics[width=0.12\linewidth]{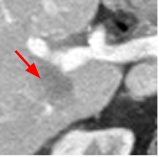}}}
            \end{overpic}
        }\hfill
        \subfloat[]{%
            \begin{overpic}[width=0.18\textwidth]{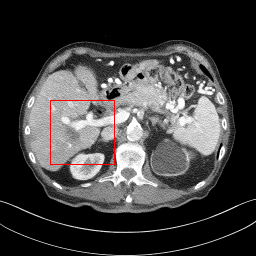}
                \put(55,-6){\adjustbox{frame=1pt,bgcolor=white}{\includegraphics[width=0.12\linewidth]{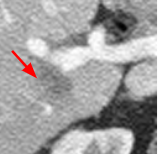}}}
            \end{overpic}
        }\hfill
        \subfloat[]{%
            \begin{overpic}[width=0.18\textwidth]{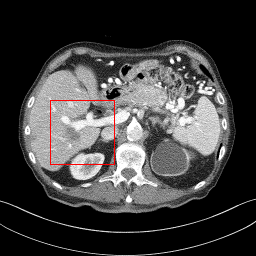}
      
                \put(55,-6){\adjustbox{frame=1pt,bgcolor=white}{\includegraphics[width=0.12\linewidth]{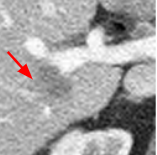}}}
            \end{overpic}
        }
    \end{minipage}
    \caption{Reconstructed LDCT images with their corresponding regions of interest. (a) Input LDCT, (b) Denomamba~\cite{ozturk2024denomamba}, (c) our method, and (d) HDCT image (Ground Truth).}
    \label{fig:res_qualitative}
\end{figure*}

The output of each VSSG is skip-connected to its corresponding input to preserve low-frequency residual information.
\subsection{Deep-Perceptual Relevancy Loss Function (DPRLF)}
We introduce, for the first time, a novel approach to the design of loss functions for perceptual learning: a Deep-Perceptual Relevancy Loss Function (DPRLF) that incorporates human visual contrast sensitivity \cite{daly1992visible,beghdadi2020critical} and information processing at multiple resolutions. Authors in a prior work~\cite{pihlgren2023systematic} investigated how feature extraction layers and backbone architectures affect the performance of perceptual loss functions. For instance, they empirically combine feature embeddings from different layers of a pretrained VGG model, but without any consideration of perceptually relevance of the extracted features to the HVS. 

The HVS sensitivity to contrast varies across spatial frequencies, achieving peak sensitivity at mid-level frequencies corresponding to textures, contours, and edges. This characteristic of the HVS is reflected through the frequency response at the contrast threshold, i.e. The Contrast Sensitivity
Function (CSF). The shape of the CSF highlights the bandpass effect of the HVS,i.e, suppressing low-contrast changes in low and high frequency bands while highlighting mid-frequency components. 

In this work, we explicitly incorporate the CSF into the computation of VGG-based perceptual embeddings. To this end, we propose a novel weighting strategy that assigns importance to different feature levels based on their perceptual relevance. Specifically, we give: i) higher weights to features extracted from mid-level layers, which capture textures, edges and other salient features; ii) moderate weights to low-level features, which represent uniform zones and low-frequency components ; and iii) lower weights to high-level semantic features (high-frequency components), which are  less relevant from a perceptual standpoint. Our DPRLF is defined as:

\begin{equation}
\mathcal{L}_{\text{DPRLF}} = \lambda_{\text{low}} \cdot \mathcal{L}_{\text{low}} + 
                              \lambda_{\text{mid}} \cdot \mathcal{L}_{\text{mid}} + 
                              \lambda_{\text{high}} \cdot \mathcal{L}_{\text{high}},
\end{equation}

where each component loss is computed as the L2 distance between the predicted and ground-truth features extracted from specific layers of a pretrained VGG16 network:

\begin{equation}
\mathcal{L}_{*} = \left\| \phi_{*}(I_{\text{pred}}) - \phi_{*}(I_{\text{gt}}) \right\|_2^2,
\end{equation}

with $* \in \{\text{low}, \text{mid}, \text{high}\}$ corresponding respectively to:
\begin{itemize}
  \item \textbf{Low-level features}: $\lambda_{\text{low}} = 0.35$,
  \item \textbf{Mid-level features}: $\lambda_{\text{mid}} = 0.5$,
  \item \textbf{High-level features}: $\lambda_{\text{high}} = 0.15$.
\end{itemize}
These weights reflect the HVS's varying sensitivity to spatial frequency bands as evidenced by the exprimentally measured CSF \cite{barten1999contrast}.



\section{Performance Evaluation}
\begin{figure*}[t]
    \centering

    \makebox[\textwidth][c]{%
    \begin{minipage}[b]{0.16\textwidth}
        \centering\begin{overpic}
            [width=\linewidth]{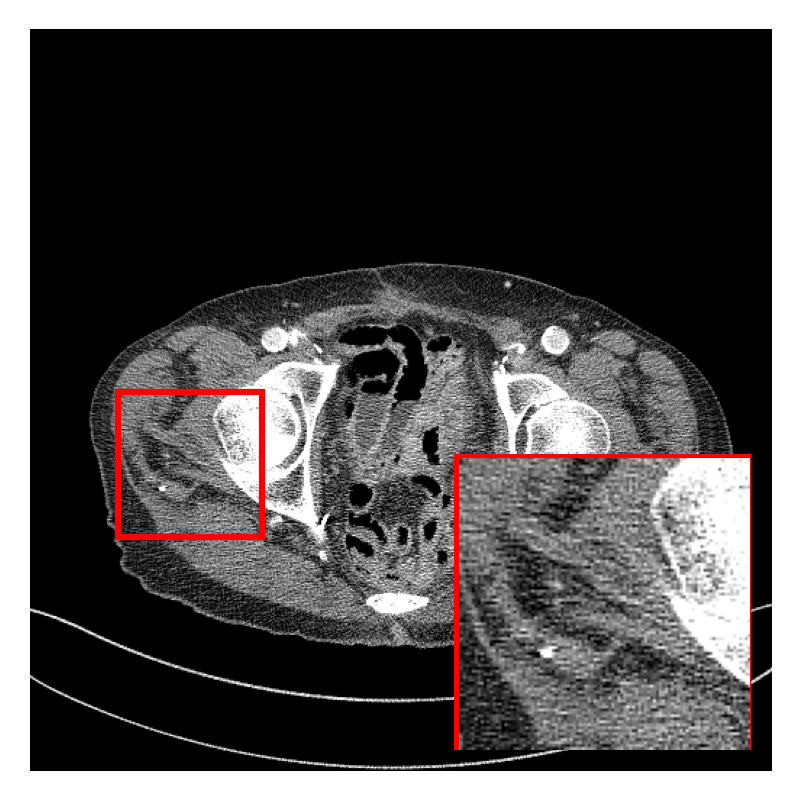}
        \put(9,80){\color{white}\textbf{(a)}}
            \end{overpic}
    \end{minipage}
    \hspace{-0.9em}
    \begin{minipage}[b]{0.16\textwidth}
        \centering\begin{overpic}[width=\linewidth]{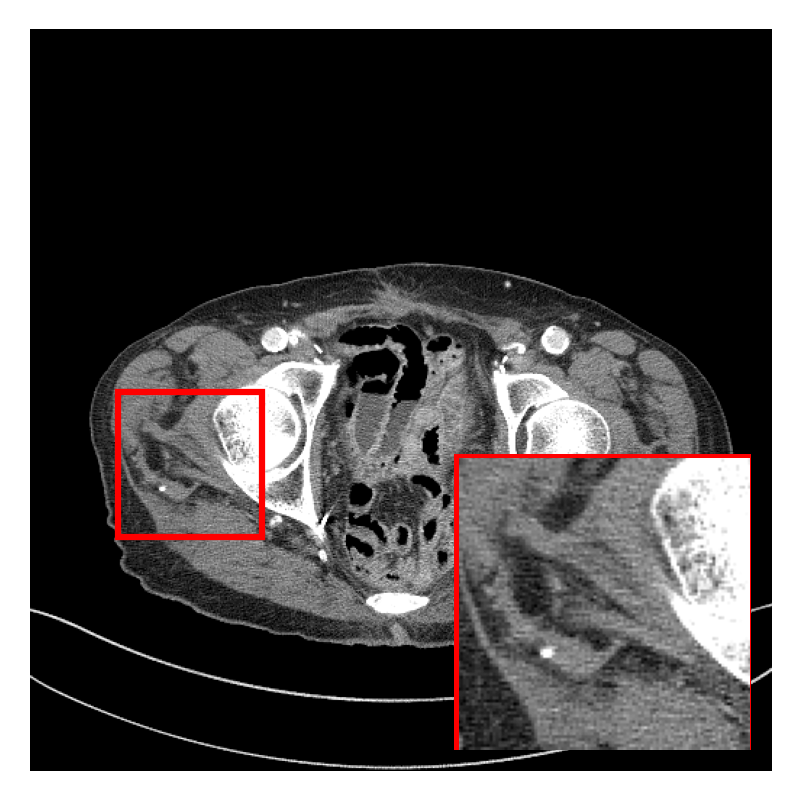}
        \put(9,80){\color{white}\textbf{(b)}}
            \end{overpic}        
    \end{minipage}
    \hspace{-0.9em}
    \begin{minipage}[b]{0.16\textwidth}
        \centering\begin{overpic}[width=\linewidth]{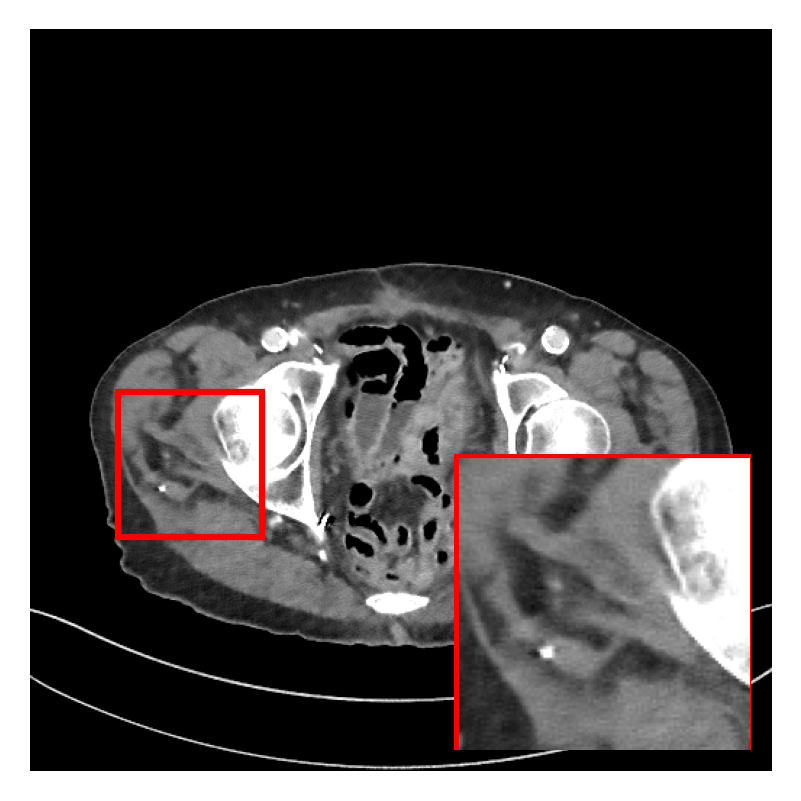}
        \put(9,80){\color{white}\textbf{(c)}}
            \end{overpic}
    \end{minipage}
    \hspace{-0.9em}
    \begin{minipage}[b]{0.16\textwidth}
        \centering\begin{overpic}[width=\linewidth]{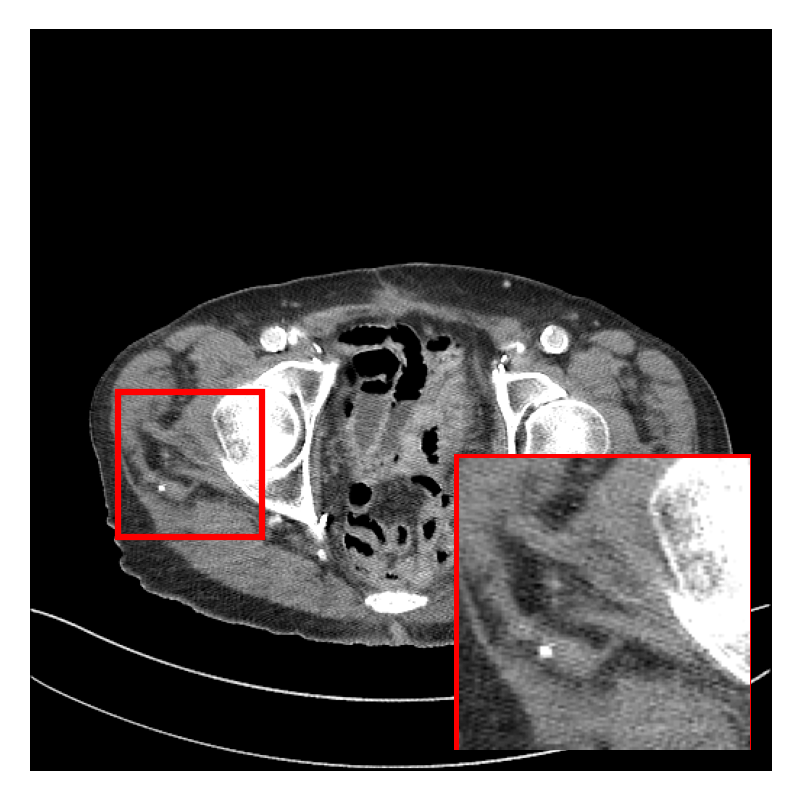}
        \put(9,80){\color{white}\textbf{(d)}}
            \end{overpic}
    \end{minipage}
    \hspace{-0.9em}
    \begin{minipage}[b]{0.16\textwidth}
        \centering\begin{overpic}[width=\linewidth]{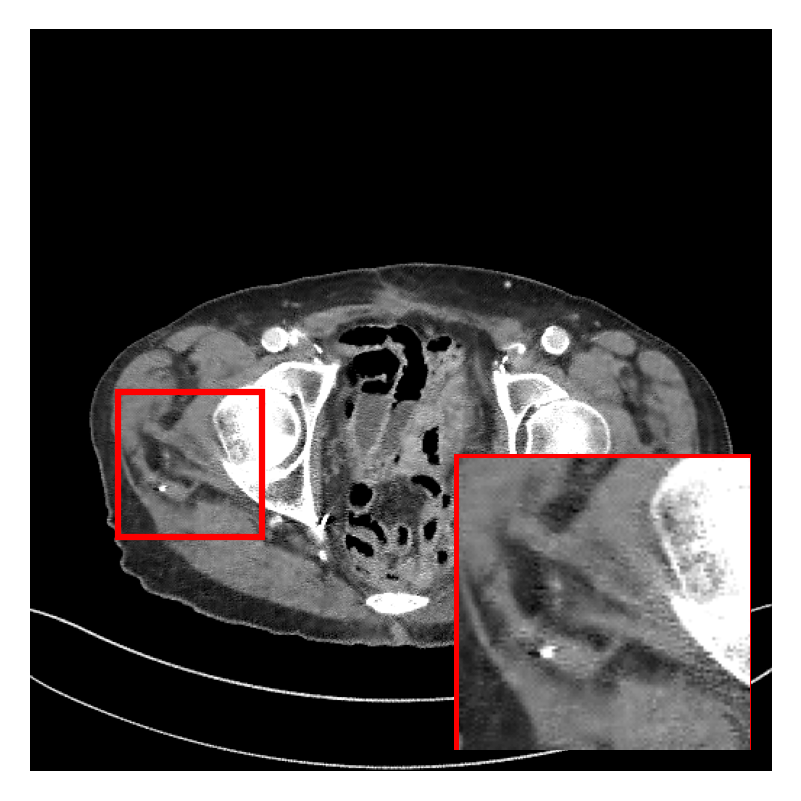}
        \put(9,80){\color{white}\textbf{(e)}}
            \end{overpic}
    \end{minipage}
    \hspace{-0.9em}
    \begin{minipage}[b]{0.16\textwidth}
        \centering\begin{overpic}[width=\linewidth]{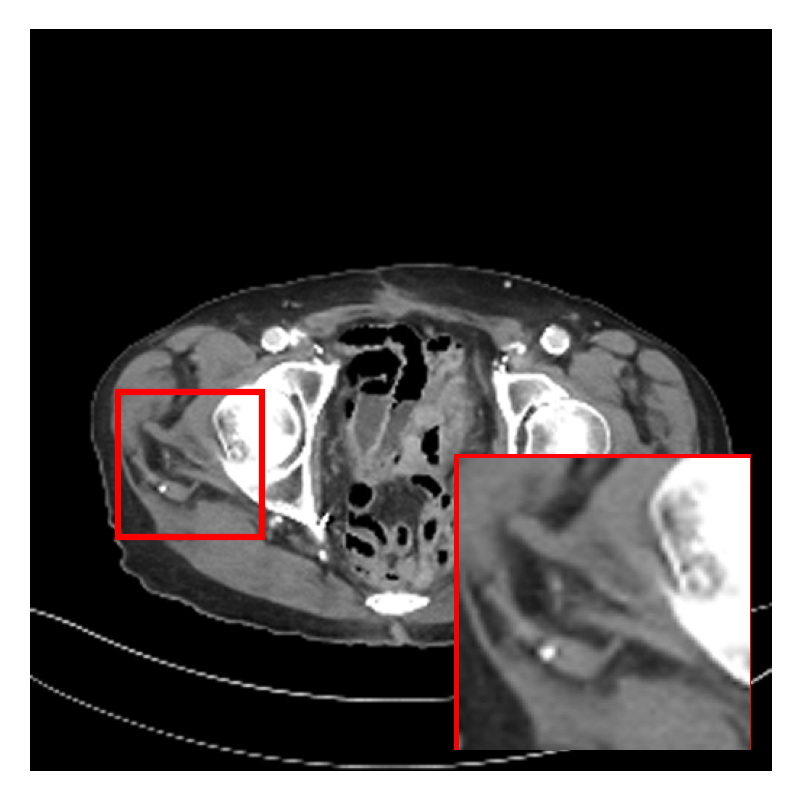}
        \put(9,80){\color{white}\textbf{(f)}}
            \end{overpic}
    \end{minipage}
    \hspace{-0.9em}
    \begin{minipage}[b]{0.16\textwidth}
        \centering\begin{overpic}[width=\linewidth]{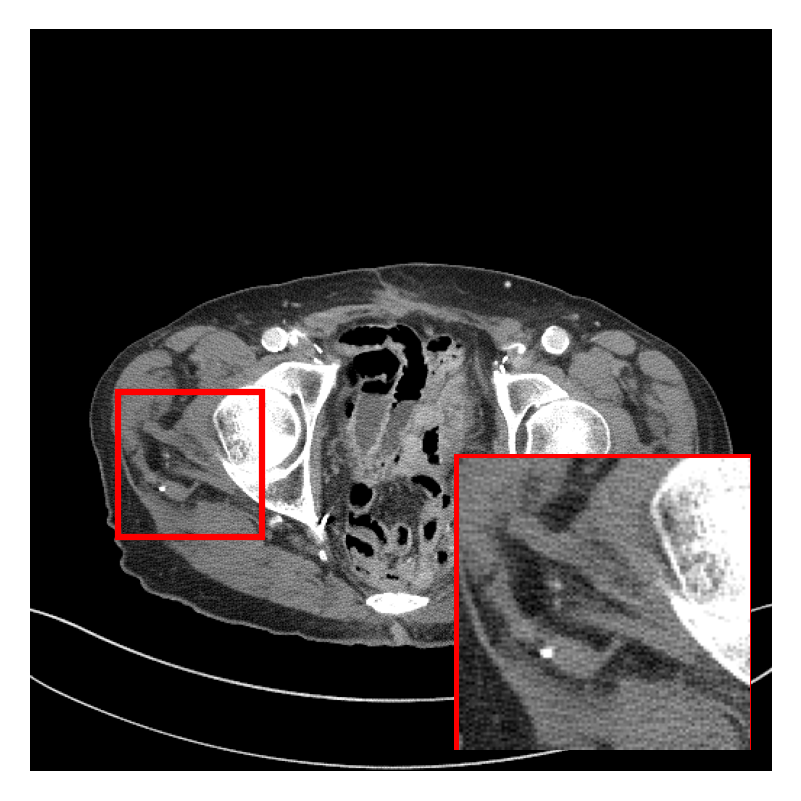}
        \put(9,80){\color{white}\textbf{(g)}}
            \end{overpic}
    \end{minipage}
    \hspace{-0.9em}
    }

    \hspace*{2em}
    \makebox[\textwidth][c]{%
    \begin{minipage}[b]{0.15\textwidth}
        \centering\includegraphics[width=\linewidth]{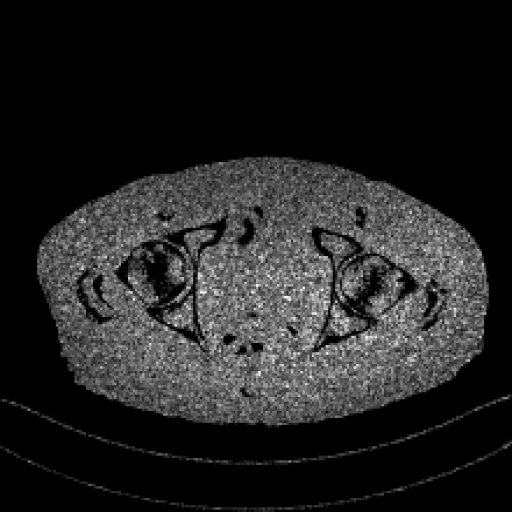}
    \end{minipage}
    \begin{minipage}[b]{0.15\textwidth}
        \centering\includegraphics[width=\linewidth]{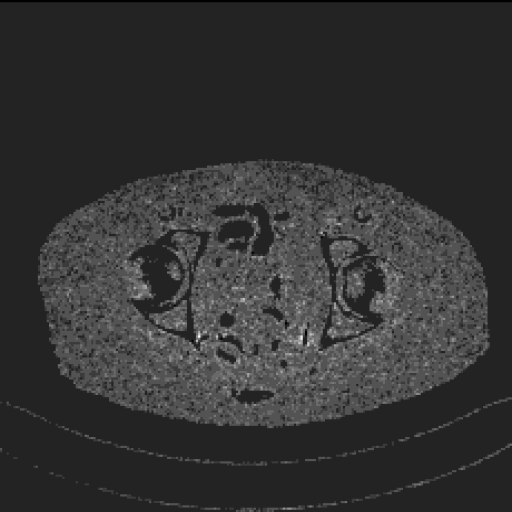}
    \end{minipage}
    \begin{minipage}[b]{0.15\textwidth}
        \centering\includegraphics[width=\linewidth]{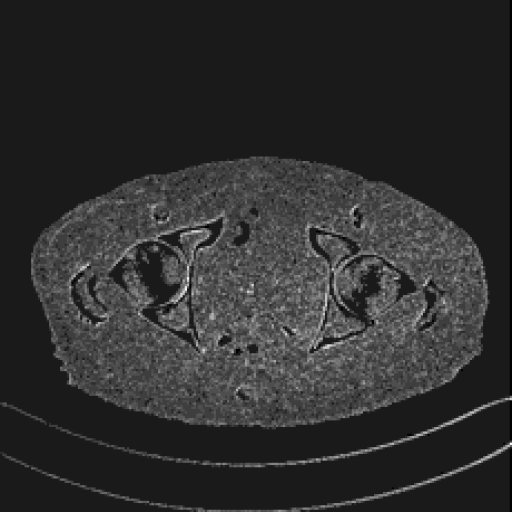}
    \end{minipage}
    \begin{minipage}[b]{0.15\textwidth}
        \centering\includegraphics[width=\linewidth]{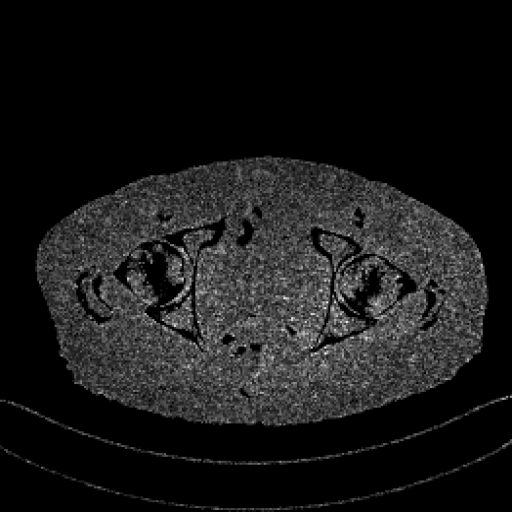}
    \end{minipage}
    \begin{minipage}[b]{0.15\textwidth}
        \centering\includegraphics[width=\linewidth]{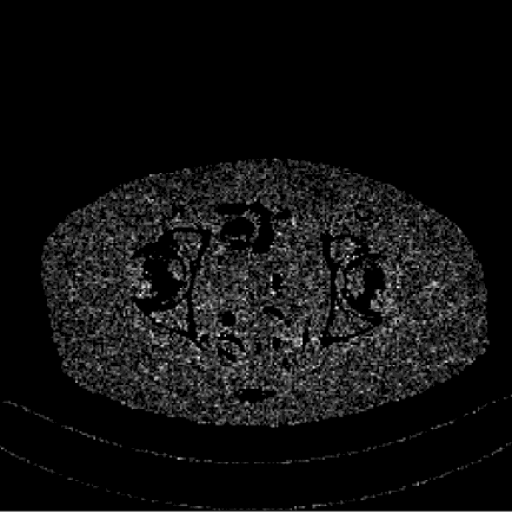}
    \end{minipage}
    \begin{minipage}[b]{0.15\textwidth}
        \centering\includegraphics[width=\linewidth]{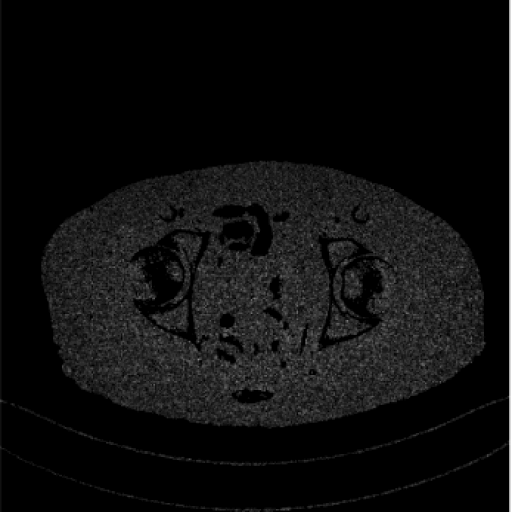}
    \end{minipage}
    \begin{minipage}[b]{0.009\textwidth} 
        \centering\includegraphics[height=2.5cm]{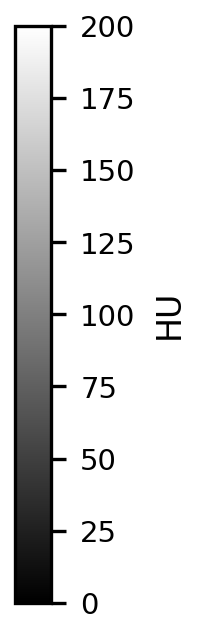}
    \end{minipage}
    }

    \caption{
    Qualitative results on slice 58 of patient L506, comparing LDCT enhancement methods.  
    (a) LDCT input; (b) HDCT (ground truth); (c)--(g) outputs of REDCNN~\cite{chen2017low}, WGAN~\cite{yang2018low}, CTFormer~\cite{wang2023ctformer}, Denomamba~\cite{ozturk2024denomamba}, and D-PerceptCT (Ours), respectively.  
    Zoomed ROIs are shown in each image. The display window is [-160, 240] HU, and absolute difference map intensities are scaled between 0 and 200 HU.
    }
    \label{fig:vis_comparison}
\end{figure*}

In this section, we first provide details of the experimental setup, including the dataset and environment employed in our study. Subsequently, we conducted an extensive performance evaluation of our proposed method, D-PerceptCT, in comparison with state-of-the-art (SOTA) approaches for LDCT enhancement. 
\subsection{Dataset, Implementation, and Evaluation}
For our study, we used the publicly available 2016 NIH-AAPM-Mayo Clinic Low Dose CT Grand Challenge dataset~\cite{mccollough2016tu}. The dataset consists of paired high-dose and quarter-dose CT images, where the high-dose CT (HDCT) images were acquired with a tube current of 200 mAs at slice thickness 3 mm. Their corresponding quarter-dose CT images were simulated by adding Gaussian-Poisson noise to HDCT projection data, based on fan-beam geometry, to mimic the degradation in image quality due to photon insufficiency when CTs are acquired at tube current around 60 mAs. The dataset has been collected from 10 patients consisting of 2,378 pairs of normal and quarter-dose 512*512 CT slices. 

The data were partitioned into training, validation, and test sets. Specifically, data from eight patients were used for training and validation, with 20\% of CT pairs are used for validation, and the model was tested on the two unseen patients.

The proposed enhancement model was trained using several widely adopted loss functions in the image quality enhancement literature, namely MSE, and Charbonnier loss functions in order to validate our proposed DPRLF. All models were trained using the Adam optimizer, employing a learning rate set to $1e-4$, and $\beta_1$ and $\beta_2$ set to 0.9 and 0.99, respectively. Training proceeded for a total of 45,000 iterations (47 epochs) with a batch size of 2, and validation was performed every 8,000 iterations. All experiments were conducted on a machine equipped with two NVIDIA A100 GPUs.

As perceptual quality is critical for diagnosis, we conducted an extensive objective evaluation using both conventional full-reference metrics (PSNR, SSIM, VIF) and perceptual metrics, including LPIPS~\cite{zhang2018unreasonable}, ST-LPIPS~\cite{ghildyal2022stlpips} and DISTS~\cite{ding2020image}. We also perform no-reference IQA using AHIQ~\cite{lao2022attentionshelpcnnsbetter}, PIQE~\cite{venkatanath2015blind} and DBCNN~\cite{Zhang_2020}, reflecting real clinical scenarios where radiologists assess image quality independently of HDCT. To better interpret performance across methods, we additionally apply a ranking strategy "beat-all-rank-first", inspired by the pairwise comparison proposed in~\cite{beghriche2025multi} (see Section 2 of the supplementary material for details).

\section{Results and discussion}
This subsection evaluates D-PerceptCT's performance against the SOTAs specifically designated for LDCT image quality enhancement  We conducted experiments and analysis, including reference-aware and blind objective quality assessment, a pair-wise comparison and residual maps analysis. Our evaluation focuses particularly on the perceptual quality of the enhanced LDCT images. Table I presents detailed full-reference and no-reference quality assessment results for D-PerceptCT alongside the competing models. D-PerceptCT demonstrates performance comparable to Denomamba [15] in traditional metrics such as PSNR, SSIM, and AHIQ. However, it significantly outperforms all the evaluated methods in providing high perceptual quality images as measured by deep perceptual quality assessment metrics LPIPS, ST-LPIPS, DISTS and PIQE. D-PerceptCT achieves the highest overall score with an accumulated performance index (API) of 46, indicating consistent performance across metrics. 
\begin{table}[htbp]
\scriptsize
\setlength{\tabcolsep}{3pt}
\renewcommand{\arraystretch}{1.7}
\caption{Quantitative evaluation of D-PerceptCT and other SOTA methods using Full-Reference and No-Reference IQA metrics (\textuparrow: Higher is better quality, \textdownarrow: Lower is better quality)}
\centering
\resizebox{\linewidth}{!}{
\begin{tabular}{|l|c|c|c|c|c|c}
\hline
\textbf{Metric} & \textbf{REDCNN~\cite{chen2017low}} & \textbf{WGAN~\cite{yang2018low}} & \textbf{CTFormer~\cite{wang2023ctformer}} & \textbf{Denomamba~\cite{ozturk2024denomamba}} & \textbf{D-PerceptCT (Ours)} \\
\hline
\multicolumn{6}{|c|}{\textbf{Full-Reference Metrics}} \\
\hline
PSNR (db,\textuparrow) & 40.04  & 39.06  & 42.15  & \textbf{43.82} & 42.97 \\
SSIM (\textuparrow)       & 0.9158 & 0.9352 & \textbf{0.9878} & 0.9809 & 0.9867 \\
VIF (\textuparrow)        & 0.7838 & 0.5183 & 0.9363 & \textbf{0.9403} & 0.6631 \\
LPIPS (\textdownarrow)    & 0.0597 & 0.0284 & 0.2742 & 0.0888 & \textbf{0.0104} \\
ST-LPIPS (\textdownarrow) & 0.0152 & 0.0101 & 0.1222 & 0.0227 & \textbf{0.0018} \\
DISTS (\textdownarrow)    & 0.1431 & 0.0778 & 0.1173 & 0.0871 & \textbf{0.0390} \\
\hline
\multicolumn{6}{|c|}{\textbf{No-Reference Metrics}} \\
\hline
AHIQ (\textuparrow)         & 0.4589 & 0.4678 & 0.4718 & \textbf{0.4950} & 0.4849 \\
PIQE (\textdownarrow)       & 31.77  & 23.63  & 23.64  & 18.83           & \textbf{12.80} \\
DBCNN (\textdownarrow)      & \textbf{0.3755} & 0.3914 & 0.4224 & 0.4671 & 0.3860 \\
\hline
\textbf{API Score} & 19 & 25 & 26 & 38 & \textbf{46} \\
\textbf{Ranking}   & 5\textsuperscript{th} & 4\textsuperscript{th} & 3\textsuperscript{rd} & 2\textsuperscript{nd} & \textbf{1\textsuperscript{st}} \\
\hline
\end{tabular}}
\label{tab:iqa_results}
\end{table}

Qualitative results in Figure \ref{fig:res_qualitative} further demonstrate that D-PerceptCT provides a better reconstruction of LDCT images, it suppresses noise while preserving edges and sharp transitions. Cropped ROIs are shown to highlight structural details relevant for diagnosis, showing that D-PerceptCT yields a better visibility of the tumor and nearby vessels in contrast to prior methods (see Section~1 of the supplementary for full comparison).

Additionally, Figure \ref{fig:vis_comparison} shows  results from another transverse CT slice. Compared to REDCNN, WGAN, CTFormer, and DenoMamba, D-PerceptCT produces sharper reconstructions with reduced noise and better preservation of structural boundaries, avoiding over-smoothing. The corresponding residual maps show that D-PerceptCT yields lower spatial errors, particularly around high-contrast bone boundaries. These results confirm that, unlike prior works that provide oversmoothed CT images, incorporating some of the HVS key aspects such as semantic-awareness and frequency sensitivity efficiently improve significantly the perceptual quality of the LDCT images.

\begin{table}[t]\footnotesize
\centering
\vspace{-0.5em}
\caption{Ablation study: The impact of loss function and architecture on LDCT image reconstruction (\textuparrow: Higher is better quality, \textdownarrow: lower is better).}
\label{tab:full_reference_iqa}
\scalebox{1}{
\begin{tabular}{l|cc|cc}
    \toprule[1pt]
    \multirow{2}{*}{\textbf{Metric}} 
    & \multicolumn{2}{c|}{\textbf{Baseline}} 
    & \multicolumn{2}{c}{\textbf{D-PerceptCT}} \\
    & \makecell{+ Charbonnier} & \makecell{+ MSE} & \makecell{+ MSE} & \makecell{+ DPRLF (Ours)} \\
    \midrule
    \multicolumn{5}{c}{\textbf{Full-Reference Metrics}} \\
    \midrule
    PSNR (dB,\textuparrow )    & \textbf{44.69} & 44.62 & 44.27 & 42.97 \\
    SSIM (\textuparrow)    & 0.9906 & \textbf{0.9918} & 0.9881 & 0.9867 \\
    LPIPS (\textdownarrow) & 0.0549 & 0.0600 & 0.0388 & \textbf{0.0104} \\
    VIF (\textuparrow)     & 0.8012 & \textbf{0.8044} & 0.7959 & 0.6631 \\
    DISTS (\textdownarrow) & 0.1417 & 0.1490 & 0.1081 & \textbf{0.0390} \\
    \bottomrule[1pt]
\end{tabular}}
\end{table}
\vspace{-0.3pt}
In table \ref{tab:full_reference_iqa} we investigate the effectiveness of our proposed DPRLF loss function. The results demonstrate that, in contrast to pixel-guided models, our method yields higher perceptual quality, as measured by deep perceptual IQAs, namely, LPIPS and DISTS scores, indicating higher perceptual quality enhanced LDCT images. These results confirm that DPRLF enhances perceptual quality as measured by objective metrics, while preserving competitive performance on conventional metrics. This underscores DPRLF’s ability to balance fidelity and preserve texture and fine anatomical details.
\section{Conclusion}
In conclusion, we introduced D-PerceptCT, a perceptually guided framework for low-dose CT enhancement that incorporates important aspects of the human visual system. A key contribution is the introduction of a novel Deep Perceptual Relevancy Loss function (DPRLF), which leverages contrast sensitivity to guide the enhancement process. D-PerceptCT achieves competitive results on PSNR, SSIM, and VIF, and clearly surpasses prior methods on perceptual IQA and qualitative visual comparisons. These results confirm the effectiveness of our approach.
Future research will extend this work by exploring the enhancement of CT images acquired at very low dose levels and involving radiologists for perceptual quality assessment, with adaptation of our approach to better meet expert needs and emphasize structures of clinical interest.

\bibliographystyle{IEEEtran}
\bibliography{bibliography}

\end{document}